\documentclass[11pt,a4paper]{article}
\usepackage[hyperref]{naaclhlt2019}
\usepackage{times}
\usepackage{latexsym}
\usepackage{xcolor}
\usepackage{todonotes}
\usepackage{amsmath}
\usepackage{subcaption}
\usepackage{floatrow}
\usepackage{graphicx}
\usepackage{bm}

\newif\ifpagelimit
\pagelimitfalse

\usepackage{url}

\aclfinalcopy 

\newcommand\sysname{\texttt{compare-mt}}

\title{\sysname: \\ A Tool for Holistic Comparison of Language Generation Systems}

\author{Graham Neubig, Zi-Yi Dou, Junjie Hu, Paul Michel, \\ 
        \textbf{Danish Pruthi, Xinyi Wang, John Wieting} \\
  Language Technologies Institute, Carnegie Mellon University \\
  {\tt gneubig@cs.cmu.edu}}

\date{}

\begin{document}
\maketitle
\begin{abstract}
In this paper, we describe \sysname, a tool for \emph{holistic} analysis and comparison of the results of systems for language generation tasks such as machine translation.
The main goal of the tool is to give the user a high-level and coherent view of the salient differences between systems that can then be used to guide further analysis or system improvement.
It implements a number of tools to do so, such as analysis of accuracy of generation of particular types of words, bucketed histograms of sentence accuracies or counts based on salient characteristics, and extraction of characteristic $n$-grams for each system.
It also has a number of advanced features such as use of linguistic labels, source side data, or comparison of log likelihoods for probabilistic models, and also aims to be easily extensible by users to new types of analysis.
\sysname{} is a pure-Python open source package,\footnote{Code \url{http://github.com/neulab/compare-mt} and video demo \url{https://youtu.be/K-MNPOGKnDQ} are available.} that has already proven useful to generate analyses that have been used in our published papers.
\end{abstract}

\section{Introduction}

Tasks involving the generation of natural language are ubiquitous in NLP, including machine translation (MT; \newcite{koehn10smt}), language generation from structured data \cite{reiter2000building}, summarization \cite{mani1999summarization}, dialog response generation \cite{oh2000stochastic}, image captioning \cite{mitchell2012midge}.
Unlike tasks that involve prediction of a single label such as text classification, natural language texts are nuanced, and there are not clear yes/no distinctions about whether outputs are correct or not.
Evaluation measures such as BLEU \cite{papineni02bleu}, ROUGE \cite{lin2014rouge}, METEOR \cite{denkowski11meteor1.3}, and many others attempt to give an overall idea of system performance, and technical research often attempts to improve accuracy according to these metrics.

However, as useful as these metrics are, they are often opaque: if we see, for example, that an MT model has achieved a gain in one BLEU point, this does not tell us what characteristics of the output have changed.
Without fine-grained analysis, readers of research papers, or even the writers themselves can be left scratching their heads asking ``\emph{what exactly is the source of the gains in accuracy that we're seeing?}''

\begin{figure}[!t]
\centering
\includegraphics[width=.9\textwidth]{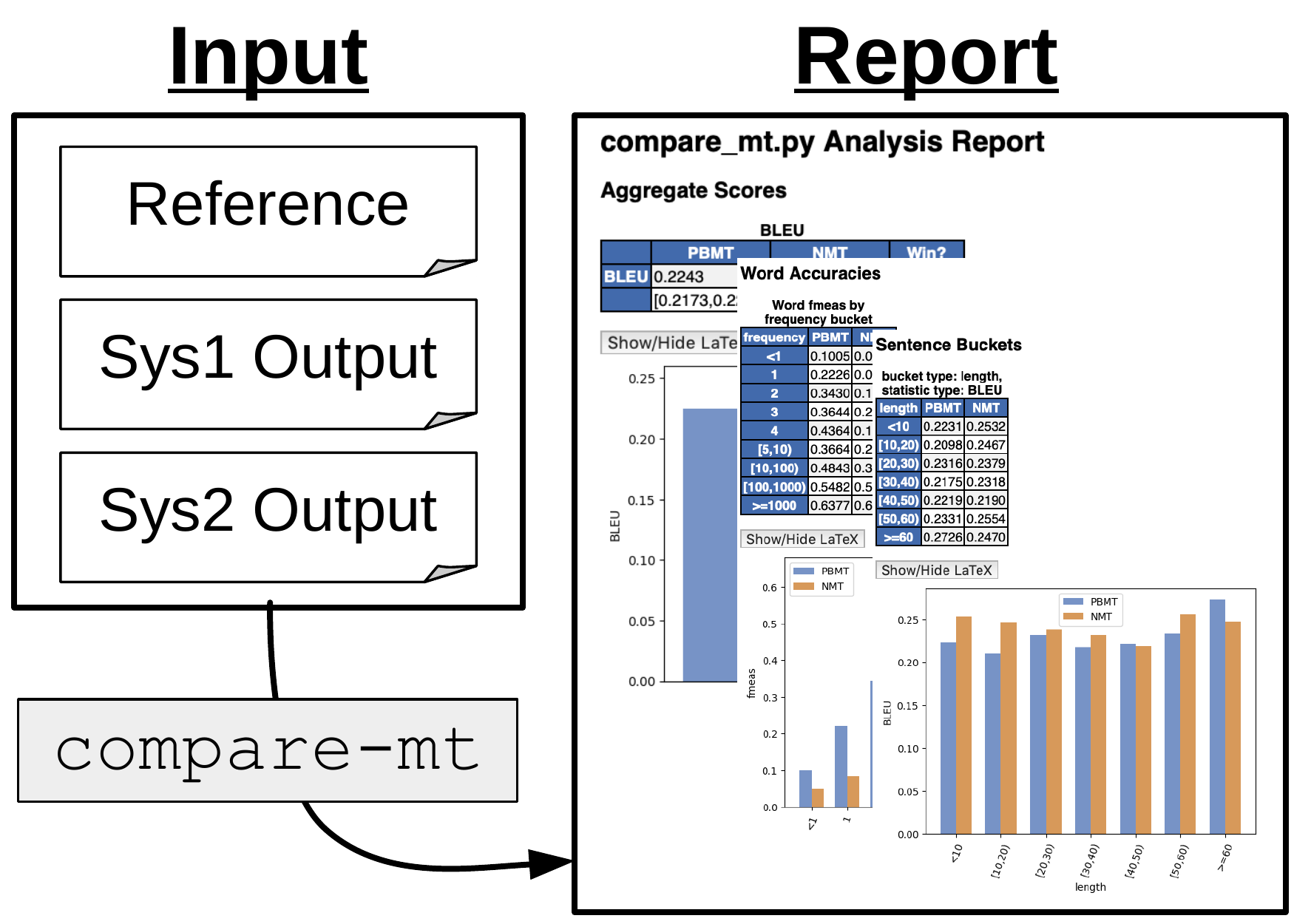}
\caption{Workflow of using \sysname{} for analysis of two systems}
\label{fig:overview} 
\end{figure}

Unfortunately, this analysis can be time-consuming and difficult.
Manual inspection of individual examples can be informative, but finding salient patterns for unusual phenomena requires perusing a large number of examples.
There is also a risk that confirmation bias will simply affirm pre-existing assumptions.
If a developer has some hypothesis about specifically what phenomena their method should be helping with, they can develop scripts to automatically test these assumptions.
However, this requires deep intuitions with respect to what changes to expect in advance, which cannot be taken for granted in beginning researchers or others not intimately familiar with the task at hand.
In addition, creation of special-purpose one-off analysis scripts is time-consuming.

In this paper, we present \sysname, a tool for \emph{holistic comparison and analysis} of the results of language generation systems.
The main use case of \sysname, illustrated in \ref{fig:overview}, is that once a developer obtains multiple system outputs (e.g. from a baseline system and improved system), they feed these outputs along with a reference output into \sysname, which extracts aggregate statistics comparing various aspects of these outputs.
The developer can then quickly browse through this holistic report and note salient differences between the systems, which will then guide fine-grained analysis of specific examples that elucidate exactly what is changing between the two systems.

Examples of the aggregate statistics generated by \sysname{} are shown in \S\ref{sec:basic}, along with description of how these lead to discovery of salient differences between systems.
These statistics include word-level accuracies for words of different types, sentence-level accuracies or counts for sentences of different types, and salient $n$-grams or sentences where one system does better than the other.
\S\ref{sec:examples} demonstrates \sysname's practical applicability by showing some case studies where has \emph{already} been used for analysis in our previously published work.
\ifpagelimit
\else
\S\ref{sec:advanced}
\fi
further details more advanced functionality of \sysname{} that can make use of specific labels, perform analysis over source side text through alignments, and allow simple extension to new types of analysis.
The methodology in \sysname{} is inspired by several previous works on \emph{automatic error analysis} \cite{popovic2011towards}, and we perform an extensive survey of the literature, note how many of the methods proposed in previous work can be easily realized by using functionality in \sysname{}, and detail the differences with other existing toolkits in
\ifpagelimit
\else
\S\ref{sec:related}.
\fi

\section{Basic Analysis using \sysname}
\label{sec:basic}

Using \sysname{} with the default settings is as simple as typing
\begin{equation*}
    \sysname\texttt{ ref sys1 sys2}
\end{equation*}
where \texttt{ref} is a manually curated reference file, and \texttt{sys1} and \texttt{sys2} are the outputs of two systems that we would like to compare.
These analysis results can be written to the terminal in text format, but can also be written to a formatted HTML file with charts and LaTeX tables that can be directly used in papers or reports.%
\footnote{In fact, all of the figures and tables in this paper (with the exception of Fig. \ref{fig:overview}) were generated by \sysname, and only slightly modified for formatting. An example of the command used to do so is shown in the Appendix.}

In this section, we demonstrate the types of analysis that are provided by this standard usage of \sysname.
Specifically, we use the example of comparing phrase-based \cite{koehn03phrasebased} and neural \cite{bahdanau15alignandtranslate} Slovak-English machine translation systems from \newcite{neubig18emnlp}.

\begin{table}[t]
  \centering
  \resizebox{\columnwidth}{!}{
  \begin{tabular}{c||ccc}
 & PBMT & NMT & Win? \\ \hline \hline
BLEU & 22.43 & \textbf{24.03} & s2$>$s1 \\
 & [21.76,23.19] & [23.33,24.65] & p$<$0.001 \\ \hline
 RIBES & 80.00 & 80.00 & - \\
 & [79.39,80.64] & [79.44,80.92] & p=0.44 \\ \hline
 Length & \textbf{94.79} & 93.82 & s1$>$s2 \\
 & [94.10,95.49] & [92.90,94.85] & p$<$0.001 \\
  \end{tabular}
  }
  \caption{Aggregate score analysis with scores, confidence intervals, and pairwise significance tests.}
  \label{tab:aggregate}
\end{table}

\ifpagelimit
\else

\begin{figure}[t]
  \includegraphics[width=.9\textwidth]{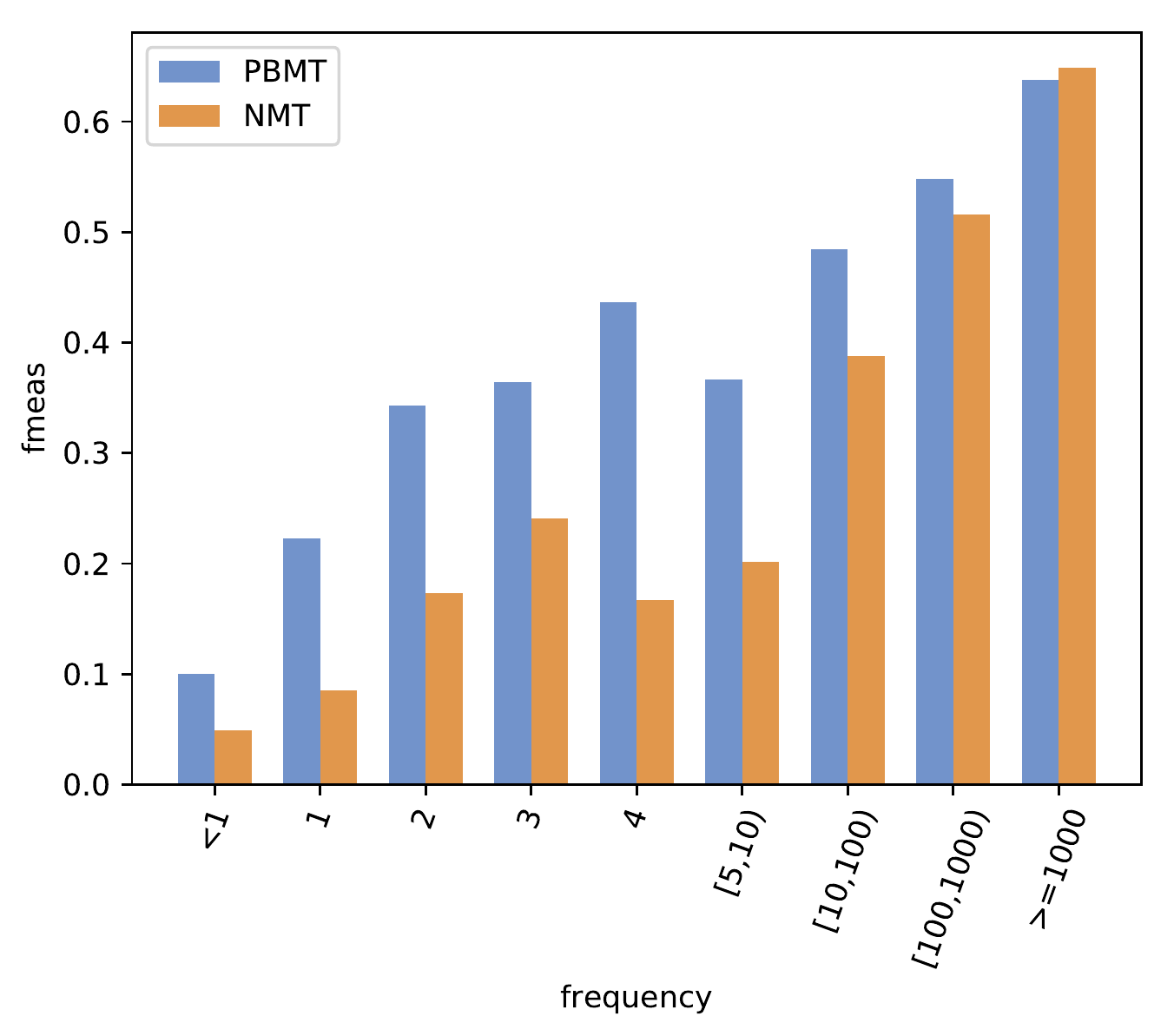}
  \caption{Analysis of word F-measure bucketed by frequency in the training set.}
  \label{fig:wordaccuracyfreq} 
\end{figure}

\begin{figure}[t]
\centering
  \includegraphics[width=.9\textwidth]{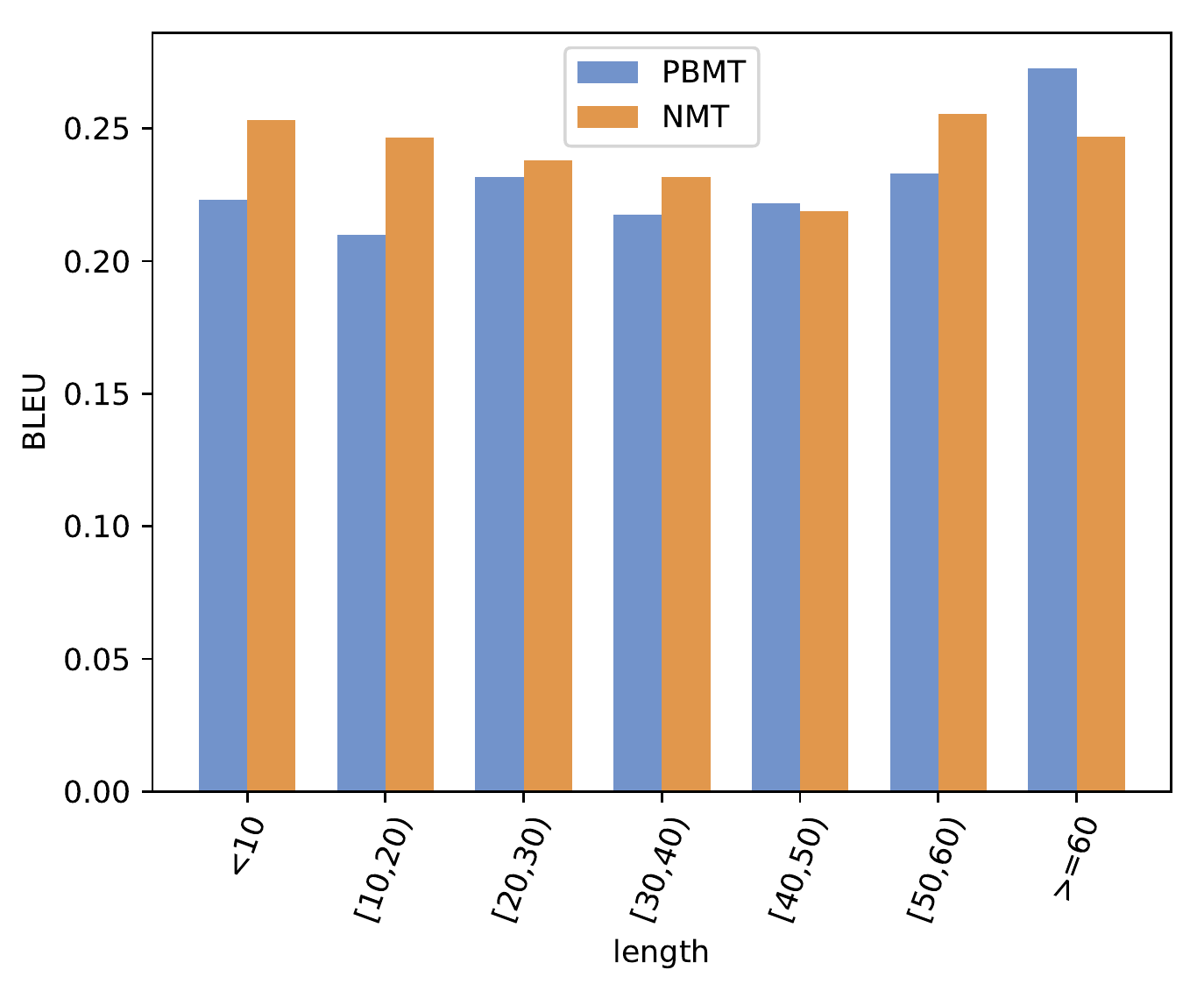}
  \caption{BLEU scores bucketed by sentence length.}
  \label{fig:sentencelengthscore} 
\end{figure}

\begin{figure}[t]
\centering
  \includegraphics[width=.9\textwidth]{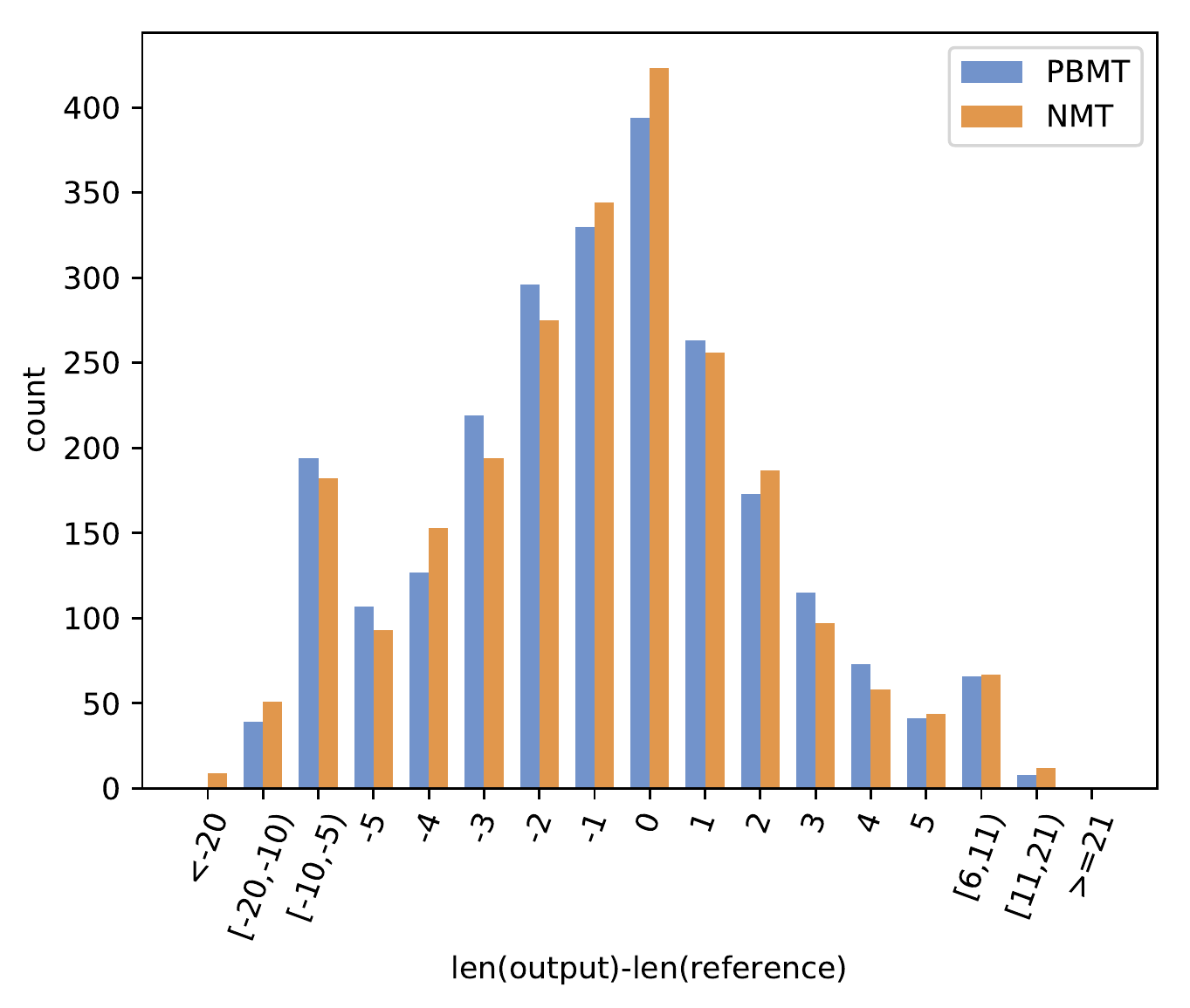}
  \caption{Counts of sentences by length difference between the reference and the output.}
  \label{fig:sentencelengthdiffcount} 
\end{figure}

\begin{figure}[t]
\centering
  \includegraphics[width=.9\textwidth]{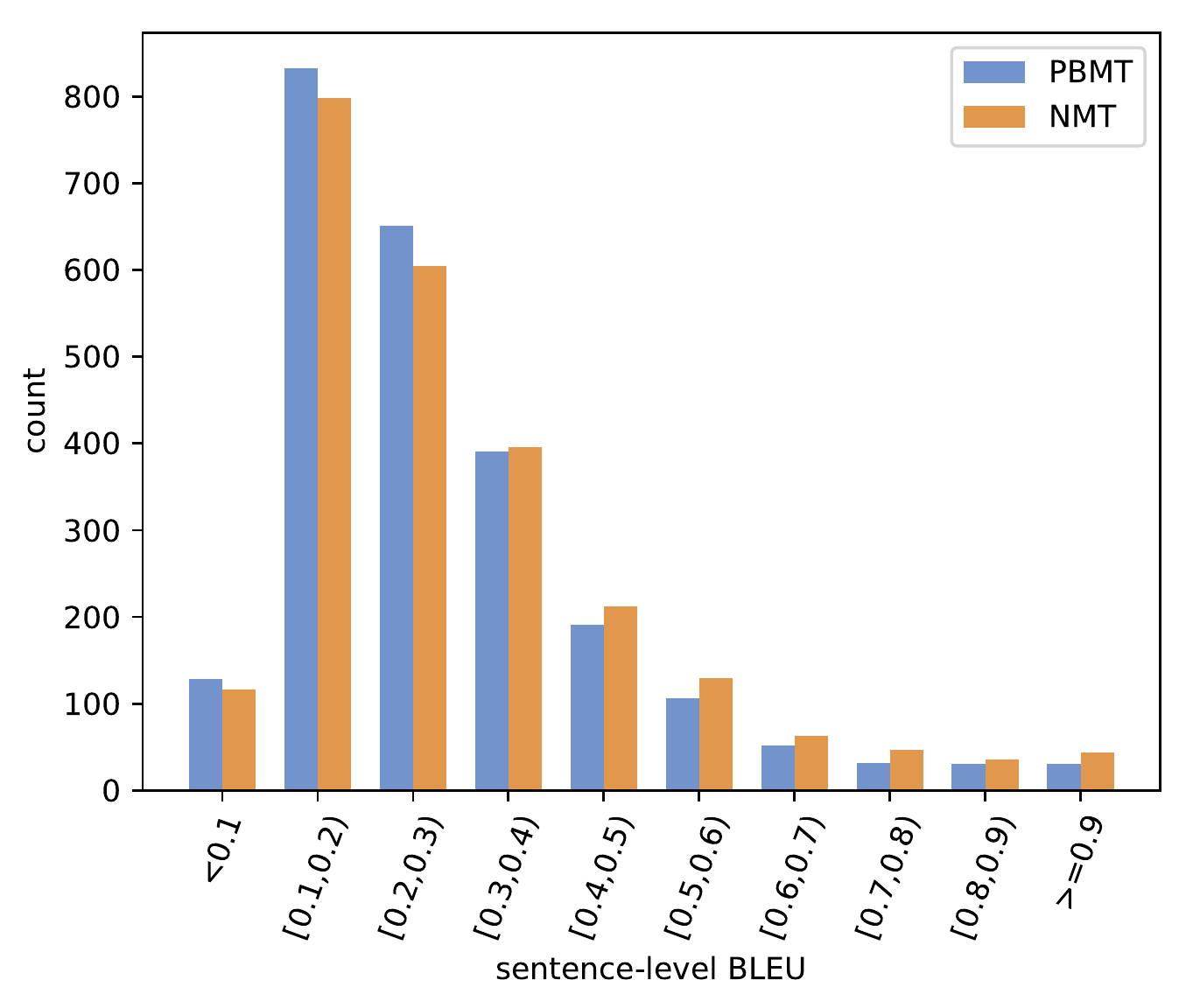}
  \caption{Counts of sentences by sentence-level BLEU bucket.}
  \label{fig:sentencesentbleucount} 
\end{figure}

\fi

\paragraph{Aggregate Score Analysis}
The first variety of analysis is not unique to \sysname, answering the standard question posed by most research papers: \emph{``given two systems, which one has better accuracy overall?''}
It can calculate scores according to standard BLEU \cite{papineni02bleu}, as well as other measures such as output-to-reference length ratio (which can discover systematic biases towards generating too-long or too-short sentences) or alternative evaluation metrics such as chrF \cite{popovic15chrf} and RIBES \cite{isozaki10ribes}.
\sysname{} also has an extensible \texttt{Scorer} class, which will be used to expand the metrics supported by \sysname{} in the future, and can be used by users to implement their own metrics as well.
Confidence intervals and significance of differences in these scores can be measured using bootstrap resampling \cite{koehn04sigtest}.

Fig. \ref{tab:aggregate} shows the concrete results of this analysis on our PBMT and NMT systems.
From the results we can see that the NMT achieves higher BLEU but shorter sentence length, while there is no significant difference in RIBES.

\paragraph{Bucketed Analysis}
A second, and more nuanced, variety of analysis supported by \sysname{} is \emph{bucketed} analysis, which assigns words or sentences to buckets, and calculates salient statistics over these buckets.

Specifically, bucketed word accuracy analysis attempts to answer the question \emph{``which types of words can each system generate better than the other?''} by calculating word accuracy by bucket.
One example of this, shown in Fig. \ref{fig:wordaccuracyfreq}, is measurement of word accuracy bucketed by frequency in the training corpus.
By default this ``accuracy'' is defined as f-measure of system outputs with respect to the reference, which gives a good overall picture of how well the system is doing, but it is also possible to separately measure precision or recall, which can demonstrate how much a system over- or under-produces words of a specific type as well.
From the results in the example, we can see that both PBMT and NMT systems do more poorly on rare words, but the PBMT system tends to be more robust to low-frequency words while the NMT system does a bit better on very high-frequency words.

A similar analysis can be done on the sentence level, attempting to answer questions of \emph{``on what types of sentences can one system perform better than the other?''}
In this analysis we define the ``bucket type'', which determines how we split sentences into bucket, and the ``statistic'' that we calculate for each of these buckets.
For example, \sysname{} calculates three types of analysis by default:
\begin{itemize}
    \item \textbf{bucket=length, statistic=score:} This calculates the BLEU score by reference sentence length, indicating whether a system does better or worse at shorter or longer sentences. From the Fig. \ref{fig:sentencelengthscore}, we can see that the PBMT system does better at very long sentences, while the NMT system does better at very short sentences.
    \item \textbf{bucket=lengthdiff, statistic=count:} This outputs a histogram of the number of sentences that have a particular length difference from the reference output. A distribution peaked around 0 indicates that a system generally matches the output length, while a flatter distribution indicates a system has trouble generating sentences of the correct length Fig. \ref{fig:sentencelengthdiffcount} indicates that while PBMT rarely generates extremely short sentences, NMT has a tendency to do so in some cases.
    \item \textbf{bucket=score, statistic=count:} This outputs a histogram of the number of sentences receiving a particular score (e.g. sentence-level BLEU score).
    This shows how many sentences of a particular accuracy each system outputs. Fig. \ref{fig:sentencesentbleucount}, we can see that the PBMT system has slightly more sentences with low scores.
\end{itemize}
These are just three examples of the many different types of sentence-level analysis that are possible with difference settings of the bucket and statistic types.

\paragraph{N-gram Difference Analysis}
The holistic analysis above is quite useful when word or sentence buckets can uncover salient accuracy differences between the systems.
However, it is also common that we may not be able to predict \emph{a-priori} what kinds of differences we might expect between two systems.
As a method for more fine-grained analysis, \sysname{} implements a method that looks at differences in the $n$-grams produced by each system, and tries to find $n$-grams that each system is better at producing than the other \cite{akabe14coling}.
Specifically, it counts the number of times each system matches each ngram $\bm{x}$, defined as $m_1(\bm{x})$ and $m_2(\bm{x})$ respectively, and calculates a smoothed probability of an $n$-gram match coming from one system or another:
\begin{equation}
    p(\bm{x}) = \frac{m_1(\bm{x}) + \alpha}{m_1(\bm{x}) + m_2(\bm{x}) + 2 \alpha}.
\end{equation}
Intuitively, $n$-grams where the first system excels will have a high value (close to 1), and when the second excels the value will be low (close to 0).
If smoothing coefficient $\alpha$ is set high, the system will prefer frequent $n$-grams with robust statistcs, and when $\alpha$ is low, the system will prefer highly characteristic $n$-grams with a high ratio of matches in one system compared to the other.

\begin{table}[t]
    \centering
    \small
    \begin{tabular}{l||ccc}
        \multicolumn{1}{c||}{\textbf{$n$-gram}} & \textbf{$m_1$} & \textbf{$m_2$} & \textbf{$s$}          \\ \hline \hline
        \texttt{phantom}           &  34 & 1 & 0.945 \\
        \texttt{Amy}               &  9  & 0 & 0.909 \\
        \texttt{, who}             &  8  & 0 & 0.900 \\
        \texttt{my mother}         &  7  & 0 & 0.889 \\
        \texttt{else happened}     &  5  & 0 & 0.857 \\ \hline
        \texttt{going to show you} &  0  & 6 & 0.125 \\
        \texttt{going to show}     &  0  & 6 & 0.125 \\
        \texttt{hemisphere}        &  0  & 5 & 0.143 \\
        \texttt{Is}                &  0  & 5 & 0.143 \\
        \texttt{'m going to show}  &  0  & 5 & 0.143 \\
    \end{tabular}
    \caption{Examples discovered by $n$-gram analysis}
    \label{tab:ngram}
\end{table}

An example of $n$-grams discovered with this analysis is shown in Tab.~\ref{tab:ngram}.
From this, we can then explore the references and outputs of each system, and figure out what phenomena resulted in these differences in $n$-gram accuracy.
For example, further analysis showed that the relatively high accuracy of ``hemisphere'' for the NMT system was due to the propensity of the PBMT system to output the mis-spelling ``hemispher,'' which it picked up from a mistaken alignment.
This may indicate the necessity to improve alignments for word stems, a problem that could not have easily been discovered from the bucketed analysis in the previous section.

\paragraph{Sentence Example Analysis}

Finally, \sysname{} makes it possible to analyze and compare individual sentence examples based on statistics, or differences of statistics.
Specifically, we can calculate a measure of accuracy of each sentence (e.g. sentence-level BLEU score), sort the sentences in the test set according to the difference in this measure, then display the examples where the difference in evaluation is largest in either direction.

\begin{table}[t]
    \centering
    \resizebox{\columnwidth}{!}{
    \begin{tabular}{c|cc}
        \textbf{Ref/Sys} & \textbf{BLEU} & \textbf{Text} \\ \hline \hline
        Ref           & -    & Beth Israel 's in Boston . \\
        \texttt{PBMT} & 1.00 & Beth Israel 's in Boston . \\
        \texttt{NMT}  & 0.41 & Beat Isaill is in Boston . \\ \hline
        Ref           & -    & And what I 'm talking about is this . \\
        \texttt{PBMT} & 0.35 & And that 's what I 'm saying is this . \\
        \texttt{NMT}  & 1.00 & And what I 'm talking about is this . \\
    \end{tabular}
    }
    \caption{Sentence-by-sentence examples}
    \label{tab:sentenceexample}
\end{table}

Tab. \ref{tab:sentenceexample} shows two examples (cherry-picked from the top 10 sentence examples due to space limitations).
We can see that in the first example, the PBMT-based system performs better on accurately translating a low-frequency named entity, while in the second example the NMT system accurately generates a multi-word expression with many frequent words.
These concrete examples can both help reinforce our understanding of the patterns found in the holistic analysis above, or uncover new examples that may lead to new methods for holistic analysis.

\begin{table*}[t]
  \centering
  \begin{tabular}{c||l}
 & Output \\ \hline \hline
Ref & And that 's me with \textbf{Youssou N 'Dour} , \textbf{onstage} , having the time of my life . \\
\texttt{PBMT} & That 's me and \textbf{Youssou N 'Dour onstage} , and he 's . \\
\texttt{NMT} & That 's me and Yosss N. \\
  \end{tabular}
  \caption{Example comparing sentences where one system did better on a particular word type}
  \label{tab:bucketedwordexample}
\end{table*}

In addition to comparing sentences where the overall translation accuracy is better or worse for a particular system, it is also possible to compare sentences where words in a particular bucket are translated more or less accuracy among the individual systems.
For example, for the ``bucketed analysis'' above, we measured the accuracy of words that appeared only one time between PBMT and NMT systems and saw that the PBMT system performed better on low-frequency words.
It is also possible to click through to individual examples, such as the one shown in Tab.~\ref{tab:bucketedwordexample}, which is an example where the PBMT system translated words in the frequency-one bucket better than the NMT system.
These examples help further increase the likelihood of obtaining insights that underlie the bucketed analysis numbers.

\section{Advanced Features}
\label{sec:advanced}

Here we discuss advanced features that allow for more sophisticated types of analysis using other sources of information than the references and system outputs themselves.

\begin{figure}[t]
  \includegraphics[width=.9\textwidth]{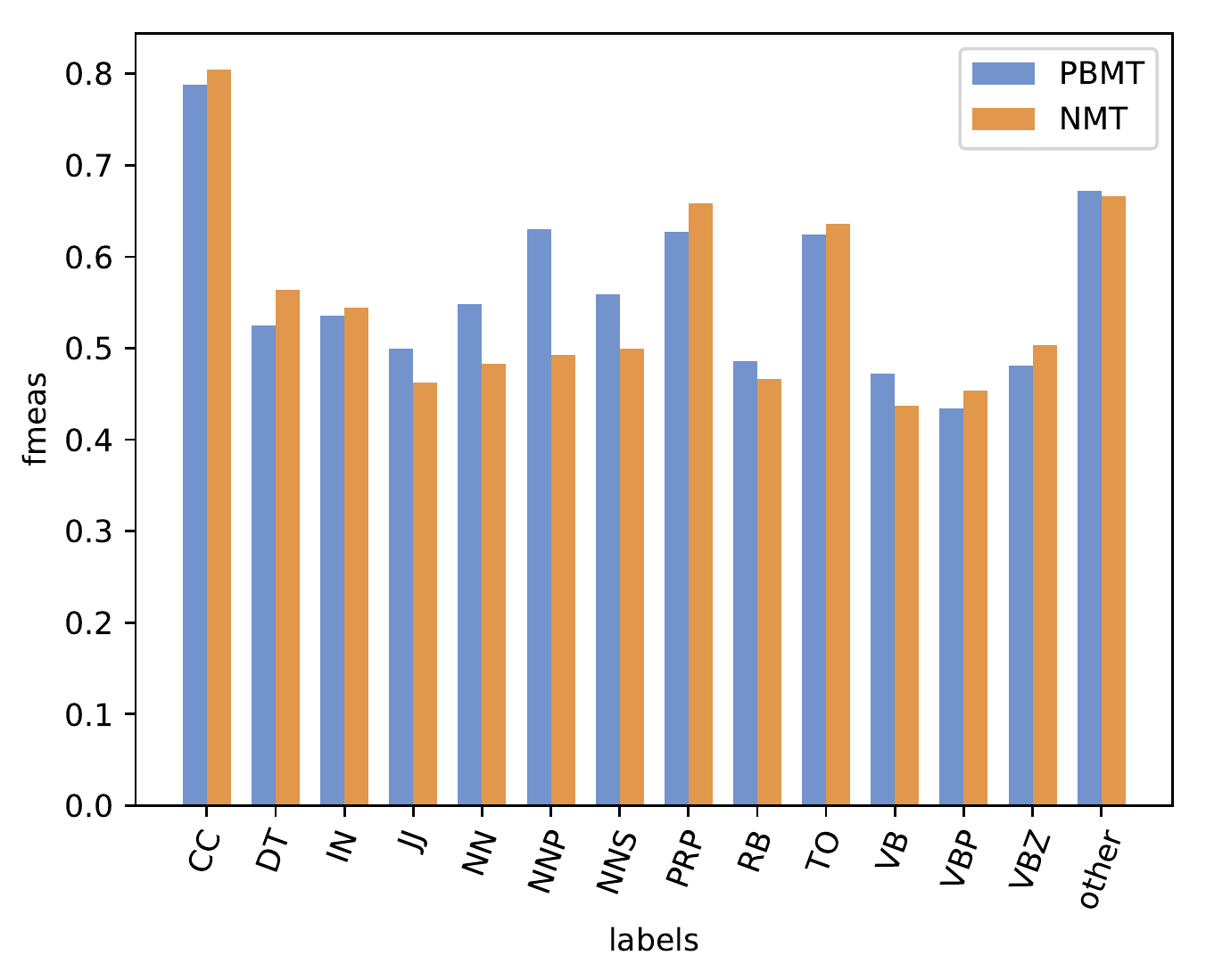}
  \caption{Word F-measure bucketed by POS tag.}
\label{fig:wordaccuracylabels} 
\end{figure}

\paragraph{Label-wise Abstraction}
One feature that greatly improves the flexibility of analysis is \sysname's ability to do analysis over arbitrary word labels.
For example, we can perform word accuracy analysis where we bucket the words by POS tags, as shown in \ref{fig:wordaccuracylabels}.
In the case of the PBMT vs. NMT analysis above, this uncovers the interesting fact that PBMT was better at generating base-form verbs, whereas NMT was better at generating conjugated verbs.
This can also be applied to the $n$-gram analysis, finding which POS $n$-grams are generated well by one system or another, a type of analysis that was performed by \newcite{chiang05hiero} to understand differences in reordering between different systems.

Labels are provided by external files, where there is one label per word in the reference and system outputs, which means that generating these labels can be an arbitrary pre-processing step performed by the user without any direct modifications to the \sysname{} code itself.
These labels do not have to be POS tags, of course, and can also be used for other kinds of analysis.
For example, one may perform analysis to find accuracy of generation of words with particular morphological tags \cite{popovic06morphosyntactic}, or words that appear in a sentiment lexicon \cite{mohammad2016translation}. 

\paragraph{Source-side Analysis}
While most analysis up until this point focused on whether a particular word on the \emph{target} side is accurate or not, it is also of interest what source-side words are or are not accurately translated.
\sysname{} also supports word accuracy analysis for source-language words given the source language input file, and alignments between the input, and both the reference and the system outputs.
Using alignments, \sysname{} finds what words on the source side were generated correctly or incorrectly on the target side, and can do aggregate word accuracy analysis, either using word frequency or labels such as POS tags.

\paragraph{Word Likelihood Analysis}
Finally, as many recent methods can directly calculate a log likelihood for each word, we also provide another tool \texttt{compare-ll} that makes it possible to perform holistic analysis of these log likelihoods.
First, the user creates a file where there is one log likelihood for each word in the reference file, and then, like the word accuracy analysis above, \texttt{compare-ll} can calculate aggregate statistics for this log likelihood based on word buckets.

\paragraph{Extending \sysname}
One other useful feature is \sysname's ability to be easily extended to new types of analysis.
For example,
\begin{itemize}
\item If a user is interested in using a different evaluation metric, they could implement a new instance of the \texttt{Scorer} class and use it for both aggregate score analysis (with significance tests), sentence bucket analysis, or sentence example analysis.
\item If a user wanted to bucket words according to a different type of statistic or feature, they could implement their own instance of a \texttt{Bucketer} class, and use this in the word accuracy analysis.
\end{itemize}

\ifpagelimit
\else
\section{Example Use-cases}
\label{sec:examples}

Over the past year or so, we have already been using \sysname{} in our research to accelerate the analysis of our results and figure out what directions are most promising to pursue next.
Accordingly, results from \sysname{} have already made it into a number of our published papers.
For example:
\begin{itemize}
    \item Figs. 4 and 5 of \newcite{wang18emnlptrdec} can be generated using sentence bucket analysis to measure ``bucket=length, statistic=score'' and ``bucket=lengthdiff, statistic=count''.
    \item Tab. 7 of \newcite{qi18naacl} shows the results of $n$-gram analysis, and Fig. 2 shows the results of frequency-based word accuracy analysis.
    \item Fig. 4 of \newcite{sachan18wmt} shows the results of frequency-based word accuracy analysis.
    \item Tab. 8 of \newcite{michel18emnlp} used \sysname{} to compare under and over-generated n-grams.
    \item Tab. 5 of \newcite{kumar2018vmf} used \sysname{} for frequency-based word accuracy analysis.
\end{itemize}

\section{Related Research and Tools}
\label{sec:related}

There have been a wide variety of tools and methods developed to perform analysis of machine translation results.
These can be broadly split into those that attempt to perform \emph{holistic analysis} and those that attempt to perform \emph{example-by-example} anaylsis.

\sysname{} is a tool for holistic analysis over the entire corpus, and many of the individual pieces of functionality provided by \sysname{} are inspired by previous works on this topic.
Our word error rate analysis is inspired by previous work on \emph{automatic error analysis}, which takes a typology of errors \cite{flanagan1994error,murata05analysis,vilar2006error}, and attempts to automatically predict which sentences contain these errors \cite{popovic2011towards,zeman2011addicter,fishel12terrorcat}.
Many of the ideas contained in these works can be used easily in \sysname{}.
Measuring word matches, insertions, and deletions decomposed over POS/morphological tags \cite{popovic06morphosyntactic,popovic-ney:2007:WMT,zeman2011addicter,elkholy11morphologicallyrich} or other ``linguistic checkpoints'' \cite{zhou08linguisticcheckpoints,naskar2011framework} can be largely implemented using the labeled bucketing functionality described in \S\ref{sec:advanced}.
Analysis of word reordering accuracy \cite{birch2010metrics,popovic2011towards,bentivogli16neuralvsphrasebased} can be done through the use of reordering-sensitive measures such as RIBES as described in \S\ref{sec:basic}.
In addition, the extraction of salient $n$-grams is inspired by similar approaches for POS $n$-gram \cite{chiang05hiero,lopez-resnik:2005:HLT-Demo} and word $n$-gram \cite{akabe14coling} based analysis respectively.
To the best of our knowledge, and somewhat surprisingly, no previous analysis tool has included the flexible sentence-bucketed analysis that is provided by \sysname{}.

One other practical advantage of \sysname{} compared to other tools is that it is publicly available under the BSD license on GitHub,\footnote{\url{https://github.com/neulab/compare-mt}} and written in modern Python, which is quickly becoming the standard program language of the research community.
Many other tools are either no longer available \cite{stymne11blast}, or written in other languages such as Perl \cite{zeman2011addicter} or Java \cite{naskar2011framework}, which provides some degree of practical barrier to their use and extension.

In contrast to the holistic methods listed above, example-by-example analysis methods attempt to intelligently visualize single translation outputs in a way that can highlight salient differences between the outputs of multiple systems, or understand the inner workings of a system.
There are a plethora of tools that attempt to make the manual analysis of individual outputs of multiple systems, through visualization or highlighting of salient parts of the output \cite{lopez-resnik:2005:HLT-Demo,stymne11blast,zeman2011addicter,madnani2011ibleu,aziz12pet,gonzalez12asiya,federmann2012appraise,chatzitheodorou13costa,klejch2015mt}.
There has also been work that attempts to analyze the internal representations or alignments of phrase-based \cite{deneefe05interactively,weese2010visualizing} and neural \cite{ding2017visualizing,lee2017visualization} machine translation systems to attempt to understand why they arrived at the decisions they did.
While these tools are informative, they play a complementary role to the \emph{holistic analysis} that \sysname{} proposes, and adding the ability to more visually examine individual examples to \sysname{} in a more extensive manner is planned as future work.

Recently, there has been a move towards creating special-purpose diagnostic test sets designed specifically to test an MT system's ability to handle a particular phenomenon.
For example, these cover things like grammatical agreement \cite{sennrich17contrastivepairs}, translation of pronouns or other discourse-sensitive phenomena \cite{muller18contextawarepronountestset,bawden18evaluatingdiscourse}, or diagnostic tests for a variety of different phenomena \cite{isabelle2017challenge}.
These sets are particularly good for evaluating long-tail phenomena that may not appear in naturalistic data, but are often limited to specific language pairs and domains.
\sysname{} takes a different approach of analyzing the results on existing test sets and attempting to extract salient phenomena, an approach that affords more flexibilty but less focus than these special-purpose methods.

\fi

\section{Conclusion}

In this paper, we presented an open-source tool for holistic analysis of the results of machine translation or other language generation systems.
It makes it possible to discover salient patterns that may help guide further analysis.

\sysname{} is evolving, and we plan to add more functionality as it becomes necessary to further understand cutting-edge techniques for MT.
One concrete future plan includes better integration with example-by-example analysis (after doing holistic analysis, clicking through to individual examples that highlight each trait), but many more improvements will be made as the need arises.

\textbf{Acknowledgements:}
The authors thank the early users of \sysname{} and anonymous reviewers for their feedback and suggestions (especially Reviewer 1, who found a mistake in a figure!).
This work is sponsored in part by Defense Advanced Research Projects Agency Information Innovation Office (I2O) Program: Low Resource Languages for Emergent Incidents (LORELEI) under Contract No. HR0011-15-C0114. The views and conclusions contained in this document are those of the authors and should not be interpreted as representing the official policies, either expressed or implied, of the U.S. Government. The U.S. Government is authorized to reproduce and distribute reprints for Government purposes notwithstanding any copyright notation here on.

\bibliography{myabbrv,gneubig}
\bibliographystyle{acl_natbib}

\clearpage

\appendix

\ifpagelimit

\fi

\begin{figure*}[t]
  \begin{verbatim}
compare-mt
  example/ted.ref.eng example/ted.sys1.eng example/ted.sys2.eng
  --compare_scores
    score_type=bleu,bootstrap=1000
    score_type=ribes,bootstrap=1000
    score_type=length,bootstrap=1000
  --compare_word_accuracies
    bucket_type=freq,freq_corpus_file=example/ted.train.eng
    bucket_type=label,ref_labels=example/ted.ref.eng.tag,out_labels=\
     "example/ted.sys1.eng.tag;example/ted.sys2.eng.tag",\
     label_set=CC+DT+IN+JJ+NN+NNP+NNS+PRP+RB+TO+VB+VBP+VBZ
  --output_directory outputs
  --sys_names PBMT NMT
  \end{verbatim}
  \caption{The command used to generate the figures and tables in this paper.}
\label{fig:examplecommand} 
\end{figure*}

\section{Example Command}
\label{sec:examplecommand}

Fig. \ref{fig:examplecommand} shows an example of the command that was used to generate the report containing the figures and tables used in this paper.

\end{document}